\documentclass[letterpaper]{article}
\usepackage[T1]{fontenc}
\usepackage[latin9]{inputenc}
\usepackage[english]{babel}
\usepackage{graphicx}
\usepackage{booktabs}
\usepackage[usenames, dvipsnames]{color}
\usepackage{epsfig}
\usepackage{arxiv}
\usepackage{amssymb}
\usepackage{epstopdf}
\usepackage{algorithm,algorithmic,amsmath}
\usepackage{subfig}
\usepackage{amsthm}
\usepackage[title]{appendix}
\usepackage{pdfpages}
\usepackage{ifthen}
\usepackage{authblk}
\usepackage{breakurl}
\newcommand{\norm}[1]{\left\lVert#1\right\rVert}

 \newlength\titlebox 
\makeatletter
\def\BState{\State\hskip-\ALG@thistlm}
\makeatother

\begin{document}
\title{\textbf{Fine-grained Uncertainty Modeling in Neural Networks}}

\author[1\thanks{Untangle AI, 79 Ayer Rajah Crescent, \#03-01, Singapore. Correspondence to: Rahul Soni \textless{}sn.rahul99@gmail.com\textgreater{}}\, ,2]{Rahul Soni}
\author[2]{Naresh Shah}
\author[2]{Jimmy D. Moore}
\affil[1]{School of Computing, National University of Singapore}
\affil[2]{Untangle AI, Singapore}
\maketitle

\begin{abstract}
Existing uncertainty modeling approaches try to detect an out-of-distribution point from the in-distribution dataset. We extend this argument to detect finer-grained uncertainty that distinguishes between (a). certain points, (b). uncertain points but within the data distribution, and (c). out-of-distribution points. Our method corrects overconfident NN decisions, detects outlier points and learns to say ``I don't know'' when uncertain about a critical point between the top two predictions. In addition, we provide a mechanism to quantify class distributions overlap in the decision manifold and investigate its implications in model interpretability.

Our method is two-step: in the first step, the proposed method builds a class distribution using Kernel Activation Vectors ($\pmb{kav}$) extracted from the Network. In the second step, the algorithm determines the confidence of a test point by a hierarchical decision rule based on the $\chi^2$ distribution of squared Mahalanobis distances.

Our method sits on top of a given Neural Network, requires a single scan of training data to estimate class distribution statistics, and is highly scalable to deep networks and wider pre-softmax layer.  As a positive side effect, our method helps to prevent adversarial attacks without requiring any additional training. It is directly achieved when the Softmax layer is substituted by our robust uncertainty layer at the evaluation phase.
\end{abstract}

\section{Introduction}
Deep Neural Networks \cite{simonyan2014very, he2016deep, szegedy2017inception} are very good estimators for prediction tasks. However, achieving a trustworthy model output has been of increasing interest recently, especially in critical areas such as autonomous vehicles or medical diagnostics. One of the ways to establish trust in the model is to be able to make its decision process interpretable \cite{zhang2018visual, samek2017explainable}. In the recent years, numerous methods \cite{ribeiro2016should,zeiler2014visualizing,kim2016examples,lundberg2017unified,koh2017understanding,biran2017human,ross2017right, springenberg2014striving, zhou2016learning, selvaraju2017grad} have been proposed to explain model predictions, for example \cite{ribeiro2016should} uses an auxiliary binary input to understand model predictions, \cite{zeiler2014visualizing} proposes inverse convolution operation using DeconvNets\cite{zeiler2011adaptive} and \cite{selvaraju2017grad} uses back-propagated gradient of the last convolution layer.

\begin{figure}[!htb]
\begin{center}
    \minipage{0.155\textwidth}
        \centering
        \includegraphics[width=\linewidth]{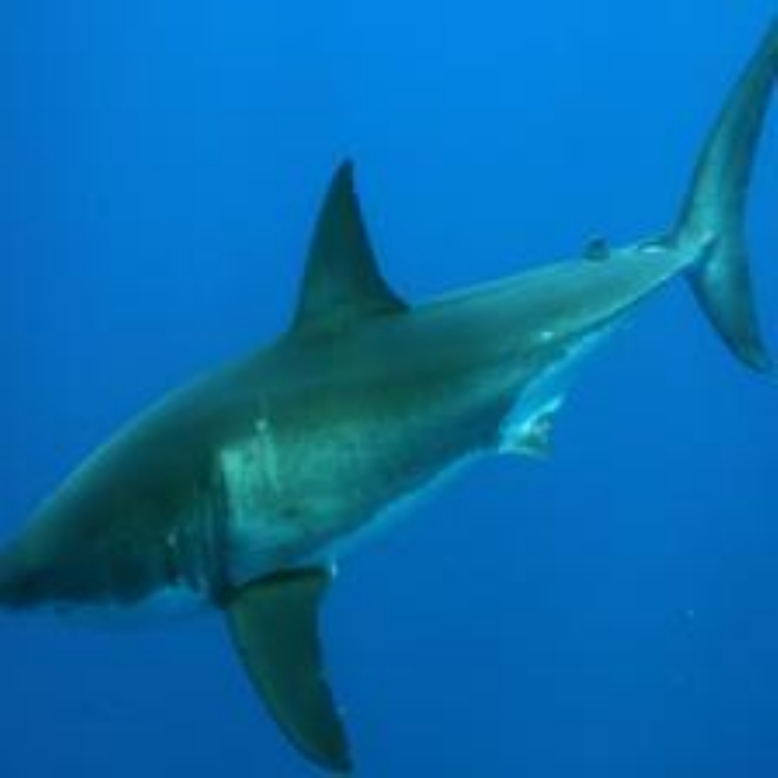}
    \endminipage\hfill
    \minipage{0.155\textwidth}
        \centering
        \includegraphics[width=\linewidth]{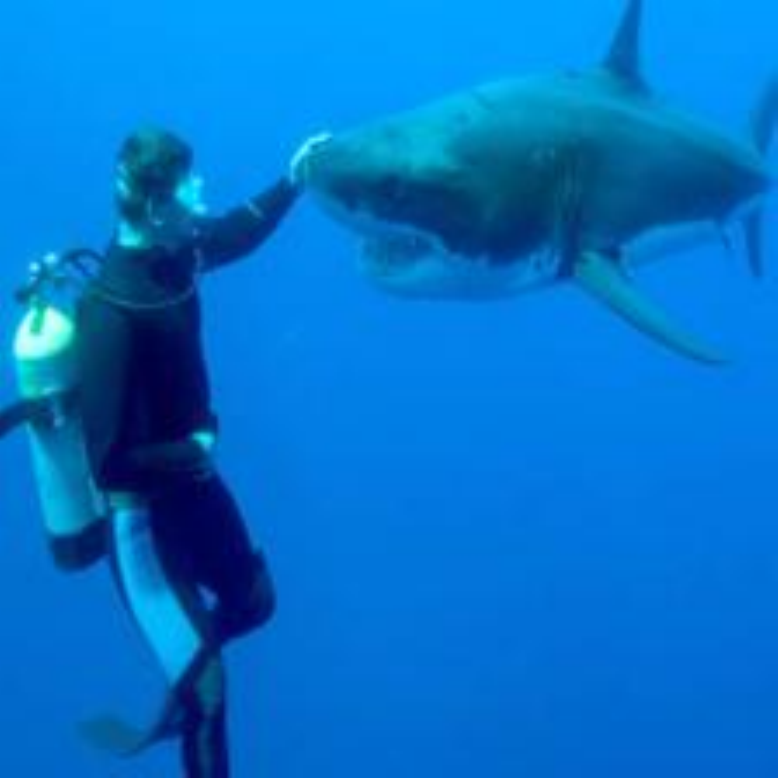}
    \endminipage\hfill
    \minipage{0.155\textwidth}%
        \centering
        \includegraphics[width=\linewidth]{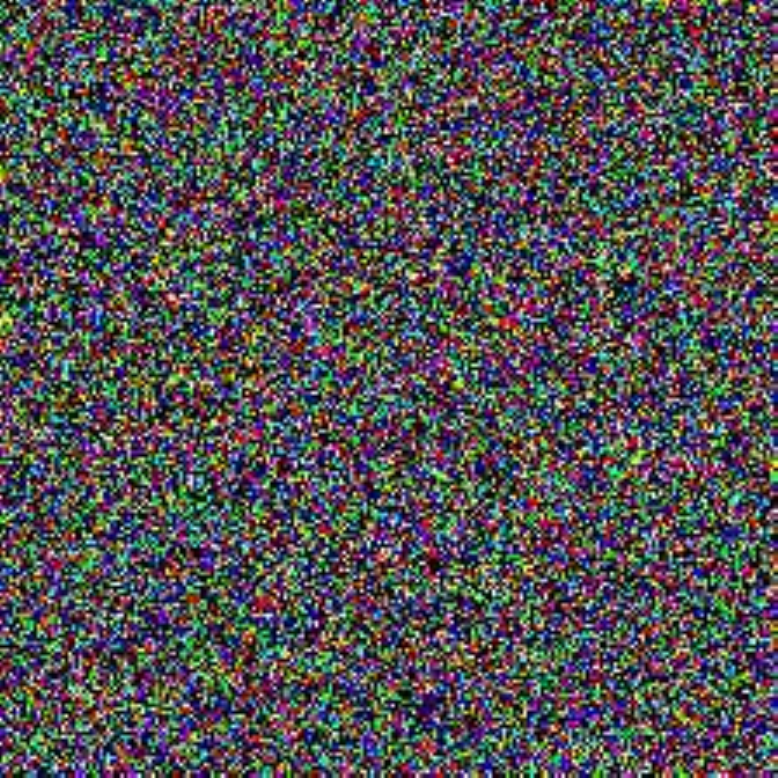}
    \endminipage
\end{center}
\begin{center}
    \minipage{0.155\textwidth}
        \centering
        \includegraphics[width=\linewidth]{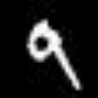}
        \captionof{subfigure}{\textit{certain}}
    \endminipage\hfill
    \minipage{0.155\textwidth}
        \centering
        \includegraphics[width=\linewidth]{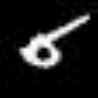}
        \captionof{subfigure}{\textit{uncertain}}
    \endminipage\hfill
    \minipage{0.155\textwidth}%
        \centering
        \includegraphics[width=\linewidth]{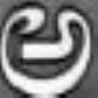}
        \captionof{subfigure}{\textit{outlier}}
    \endminipage
\caption{Prediction confidence on our method versus the Softmax and previous works \cite{lee2018simple, sensoy2018evidential, kaplan2018uncertainty, nader2014mahalanobis}. \textbf{Our method:} \textbf{(a)} (top) is detected as a \textit{certain} ``great white shark'', and (bottom) is detected as a \textit{certain} ``digit-9''; \textbf{(b)} (top) is detected as an \textit{uncertain} point between ``great white shark'' and  ``scuba diver'', and (bottom) is detected as an uncertain point between ``digit-5'' and ``digit-8''; \textbf{(c)} both (top) and (bottom) are detected as \textit{out-of-distribution} points. \textbf{Softmax and previous works}: \textbf{(a)} (top) is detected as a \textit{certain} ``great white shark'', and (bottom) is detected as a certain ``digit-9''; \textbf{(b)} and \textbf{(c)} both (top) and (bottom) are detected as the \textit{out-of-distribution} points}.
\label{fig:fig1}
\end{center}
\end{figure}

Most of the approaches in model explanations are gradient based and operate on the input without regards to the model prediction confidence. While the input explanations might be still meaningful, the Softmax still produces overconfident scores which are not inline with the explanations. For example, the VGG19 pretrained model is overconfident about top-1 prediction than the second top prediction even though both predictions have equal prior as shown in Fig.[\ref{fig:fig1}]. 

Although significant work has been done on explanations at the input level, estimating the uncertainty in prediction is still a challenge and is an active area of research \cite{feinman2017detecting, hendrycks2016baseline, liang2017enhancing, sensoy2018evidential, kaplan2018uncertainty, lee2017training}. One of the first approaches in uncertainty calibration \cite{guo2017calibration} was to scale the logit values in the Softmax (also called temperature scaling), however, temperature scaling alone is insufficient to detect outlier points on the lower end of the probabilities. \cite{feinman2017detecting} is proposed to prevent adversarial attacks by getting an estimate of the manifold for each class. \cite{hendrycks2016baseline} employs a simple mechanism to separate the test data into correctly and incorrectly classified examples and then plots an AUCROC curve to summarize the performance of a binary classifier discriminating with the score. \cite{liang2017enhancing} uses temperature scaling and adds a small amount of perturbation in the input image to separate in- and out-of-distribution samples. \cite{sensoy2018evidential} and \cite{kaplan2018uncertainty} are based on Dirichlet distribution \cite{blei2003latent} of class probabilities to form subjective opinions (however, a small change in the uncertainty threshold can significantly affect the model confidence). \cite{lee2017training} enhances the training step by adding a confidence score in the loss function to remove out-of-distribution examples. \cite{nader2014mahalanobis} only detects outlier points based on a fixed threshold. \cite{lee2018simple} is similar to our work where it uses the pre-softmax layer to estimate the prior distribution of classes and then detect out-of-distribution points by setting a threshold on Mahalanobis distances. However, this approach is not scalable to wide pre-softmax layers due to joint covariance matrix computation. Besides, the proposed threshold-based separation method is insufficient to separate uncertain points within the distribution and out-of-distribution points (thus lacking finer-grained uncertainty).

\textbf{Contribution: }\textbf{(a)} We propose a robust framework - an extension of \cite{lee2018simple} - to separate out-of-distribution points from uncertain points within the distribution, \textbf{(b)} Propose kernel activation vectors, $\pmb{kav}$, that capture the low-level and high-level activations, \textbf{(c)} Retrieve relevant training points in the active learning framework, \textbf{(d)} Propose diagonal covariance matrix assumption in the class distribution that makes the framework scalable to very deep networks and or very wide pre-softmax layers without compromise in prediction performance, \textbf{(e)} Quantify class distribution overlap to understand concept affinity and provide model interpretability, and \textbf{(f)} Define a new baseline to measure outlier-detection capacity of the model.

\section{Uncertainty Modelling}
Given a neural network pre-trained for a specific target, for example, multi-class classification, we aim to evaluate the predictive uncertainty of the top prediction. At the training step, we first model the prior distribution of each class as a multi-variate normal distribution capturing the distribution of input activations. In order to provide a general procedure, we refer to these activations as Kernel Activation Vector (kav) that captures the low-level and high-level activations of the manifold for each class. At the training step, we also compute and store the top-k nearest KAV members (k being the hyperparameter) for each training class. These member vectors are then used in the ensemble voting for uncertainty prediction as discussed later in this section. The overall training step is summarized in algorithm~\ref{alg:uncrt}. At the evaluation step, the uncertainty score is assigned based on a hierarchical decision process as outlined in detail in the following two subsections. 

%Given a neural network pretrained for a specific target, for example multi-class classification, we aim to evaluate the predictive uncertainty of the top prediction. We first define an average \textbf{Kernel Activation Vector} (KAV) - a compressed representation vector that captures the low-level and high-level activations of the manifold for each class. For the convolutional layers, KAV is formed by taking an average of each feature map in the layer; for the Dense layers and the BatchNorm layers, KAV is equal to the ReLu activated output of that layer. At uncertainty estimation step, for each target class $c_i$, every training example $\pmb{x}_i \in c_i$ is represented as a concatenation of KAVs of all kernel layers. We call it ``Composite KAV''. Similar to \cite{lee2018simple}, we first model the prior distribution of each class as a multi-variate normal distribution capturing the distribution of KAVs. At training step, we also compute and store the top-k nearest KAV members (k being the hyperparameter) for each training class. These member vectors are then used in the ensemble voting for uncertainty prediction as discussed later in this section. The overall training step is summarized in algorithm \ref{algo:algo1}. At evaluation step, the uncertainty score is assigned based on a hierarchical decision process as outlined in detail in the following two subsections.

\begin{algorithm}[!htb]
\caption{Compute statistics ($\pmb{\mu}_j$, ${\pmb{\sigma}_j}^2$) of class specific prior distribution. Compute $k$ nearest members ($C_{N_j} \forall j \in [1, \cdots, k]$) of that distribution}
   \label{alg:uncrt}
\begin{algorithmic}
   \STATE {\bfseries Input:} Training instances $\boldsymbol{x}_i \in \mathbb{R}^n$, trained Neural Network estimator, $f(\cdot)$
   \STATE Initialize $\textit{N} \gets \textit{number of training points}$
   \STATE Initialize $k \gets \textit{number of target classes}$
   \STATE Initialize $f(\cdot) \gets \textit{NN estimator}$
   \STATE Initialize $g(f(\cdot), x) \gets \textit{KAV extractor}$
   \STATE Initialize $\pmb{\mu}_j \gets \textit{zero vector} \in \mathbb{R}^d \, \, \forall \, j \in [1, \cdots, k]$
   \STATE Initialize ${\pmb{\sigma}_j}^2 \gets \textit{zero vector} \in \mathbb{R}^d \, \, \forall \, j \in [1, \cdots, k]$
   \STATE Initialize $\pmb{M}_j \gets \textit{list of nearest neighbours} \, \, \forall \, j \in [1, \cdots, k]$
   \STATE Initialize $N_j=0 \gets \textit{count of samples} \, \, \forall \, j \in [1, \cdots, k]$

\FOR{$j=1$ {\bfseries to} $k$}
    \FOR{$i=1$ {\bfseries to} $N$}
        \STATE $\pmb{x}_{ji} \gets i^{th} \textit{ training sample} \in C_j \, class$
        \STATE $\pmb{kav}_{ji} \gets g(f, \pmb{x}_{ji})$
        \STATE $\pmb{\mu}_j \gets \pmb{\mu}_j + \pmb{kav}_{ji}$
        \STATE ${\pmb{\sigma}_j}^2 \gets {\pmb{\sigma}_j}^2 + \pmb{kav}_{ji}^2$
        \STATE $N_j \gets N_j + 1$\\
    \ENDFOR
\ENDFOR

\FOR{$j=1$ {\bfseries to} $k$}
    \STATE $\pmb{\mu}_j \gets \frac{\pmb{\mu}_j}{N_j}$
    \STATE ${\pmb{\sigma}_j}^2 \gets \frac{{\pmb{\sigma}_j}^2}{N_j} - \pmb{\mu}_j^2$
\ENDFOR

\FOR{$j=1$ {\bfseries to} $k$}
    \STATE $\pmb{M}_j \gets \textit{M nearest members of } C_j$
\ENDFOR
\end{algorithmic}
\end{algorithm}

\subsection{Distance Estimates}
Given an observation point, $\pmb{x}_i$, we extract the kernel activation vector, $\pmb{kav}_i$, and compute its Mahalanobis distance from a multi-variate Gaussian distribution as:
%The proposed hierarchical decision rule is based off of Mahalanobis distances from the class specific prior distributions, joint distribution of top-2 classes, and the separation of classes in the higher dimension multivariate distribution. We first extract $\pmb{kav}$ from the pre-trained NN and define the distances as follows:
\begin{equation}\label{eq:1}
    \begin{split}
     d_{ij} &= \sqrt{\big(\pmb{kav}_i - \pmb{\mu}_j\big)^T \pmb{\Sigma}_{j}^{-1} \big(\pmb{kav}_i - \pmb{\mu}_j\big)}
    \end{split}
\end{equation}
where, $\pmb{kav}_i \in \textit{class} \,\, C_j$ is the kernel activation vector extracted from the pre-trained NN. Its belongingness to class $C_j$ is determined by the Maximum Likelihood Estimate (MLE) on pre-softmax layer of the NN esitmator. Note that computing the dense Covariance matrix, $\pmb{\Sigma}$, in a high dimensional space is an expensive operation, both in time $\left(\mathcal{O}(n^3)\right)$ and space $\left(\mathcal{O}(n^2)\right)$. We assume that the elements in $\pmb{kav}_i$ are linearly independent and uncorrelated, which reduces the Covariance matrix to a diagonal matrix. This linear assumption reduces both time and space complexity to $\left(\mathcal{O}(n)\right)$ and makes the training procedure scalable to large datasets and wider Softmax layer. The Mahalanobis distance is therefore modified as:
\begin{equation}\label{eq:2}
    \begin{split}
     d_{ij} &= \sqrt{\sum_{k=1}^{N} \frac{((\pmb{kav}_i)_k - (\pmb{\mu}_j)_k)^2}{(\sigma_{j}^2)_k}} = \norm{\Tilde{\pmb{kav}_{ij}}}\\
    \end{split}
\end{equation}
where $\Tilde{\pmb{kav}_{ij}} = \frac{\pmb{kav}_i - \pmb{\mu}_j}{\pmb{\sigma}_{j}}$ is a transformed kernel activation vector which is normalized to the $j^{th}$ class, i.e., centered by mean ($\pmb{\mu}_j$) and scaled by the standard deviations ($\pmb{\sigma}_j$) element wise. Empirically, we validate that this constraint still gives superior results compared to baseline \cite{hendrycks2016baseline} and previous works \cite{lee2018simple, liang2017enhancing}.

We also define an overlap between prior distributions of different classes using Bhattacharya distances to better understand the decision manifold. Bhattacharya distance, $d_{jk}$, between classes $C_j$ and $C_k$, is given as:
\begin{equation}\label{eq:3}
    \begin{split}
    d_{jk} &= \frac{1}{8} \big(\pmb{\mu_1} - \pmb{\mu_2}\big)^{T} \pmb{\Sigma}^{-1} \big(\pmb{\mu_1} - \pmb{\mu_2}\big) + \frac{1}{2}\ln{\frac{|\pmb{\Sigma}|}{\sqrt{|\pmb{\Sigma}_1||\pmb{\Sigma}_2|}}}\\
    \end{split}
\end{equation}
Since the length of $\pmb{kav}$ scales with the depth of a CNN, the second term in Eq.[\ref{eq:3}] becomes numerically unstable at large dimensions, for example, VGG16 produces $\pmb{kav} \in \mathbb{R}^{13614}$. We use the following optimizations in Eq.[\ref{eq:3}] to achieve numerical stability:
\begin{equation}\label{eq:4}
    \begin{split}
    &\frac{1}{2}\ln{\frac{|\pmb{\Sigma}|}{\sqrt{|\pmb{\Sigma}_1||\pmb{\Sigma}_2|}}} = \frac{1}{2}\ln{|\pmb{\Sigma}|} - \frac{1}{4} \ln{|\pmb{\Sigma}_1|} - \frac{1}{4} \ln{|\pmb{\Sigma}_2|}\\
    &= \frac{1}{2}\ln{\Bigg(\prod_{i=1}^N {\sigma^2}_i\Bigg)} - \frac{1}{4} \ln{\Bigg(\prod_{i=1}^N {\sigma^2}_{1i}\Bigg)} - \frac{1}{4} \ln{\Bigg(\prod_{i=1}^N {\sigma^2}_{2i}\Bigg)}\\
    &= \frac{1}{2} \sum_{i=1}^N \ln{\big({\sigma^2}_i\big)} - \frac{1}{4} \sum_{i=1}^N \ln{\big({\sigma^2}_{1i}\big)}  - \frac{1}{4} \sum_{i=1}^N \ln{\big({\sigma^2}_{2i}\big)}\\
    &= \frac{1}{4} \sum_{i=1}^N \left( 2\ln{\left( \frac{{\sigma^2}_{1i} + {\sigma^2}_{2i}}{2} \right)} - \ln{\big({\sigma^2}_{1i}\big)} - \ln{\big({\sigma^2}_{2i}\big)} \right)
    \end{split}
\end{equation}
where, the second step uses the proposed uncorrelation assumption (diagonal Covariance matrix). We now define the hierarchical uncertainty decision process.

\subsection{Hierarchical Uncertainty Modelling:}
For every test point, the proposed approach computes the Mahalanobis distance (Eq.[\ref{eq:2}]) of the kernel activation vector $\pmb{kav}$ from the distributions of top-2 classes. A hierarchical decision function takes these distances and categorizes the test point as \textit{outlier}, \textit{uncertain}, or \textit{certain}.

We define a point as out-of-distribution point if it is outside of the $1^{st}-\textit{quantile}$ of the distribution. To determine the $1^{st}-\textit{quantile}$ range, or, in general, the $k^{th}-\textit{quantile}$ range, we note that the squared Mahalanobis distances follow ${\chi^2}_{df}$ distribution \cite{hardin2005distribution} whose degree of freedom $df$ is equal to the dimension of the distribution. Since the ${\chi^2}_{df}$ distribution does not have a closed from Cumulative Distribution Function (CDF), determining its $k^{th}-\textit{quantile}$ although possible under approximations \cite{okagbue2017quantile}, is still a complicated procedure. We observe that, at very large degree of freedom, the ${\chi^2}_{df}$ can be approximated as a normal distribution with $\mu = df$ and $\sigma = 2\cdot df$. Let us denote this \textit{distance distribution} as $N_{MH}(df, 2df)$ to distinguish it from the \textit{activation distributions} (of $\pmb{kav}$).

We use the out-of-distribution criteria described above to achieve finer grained uncertainty - separation of outlier points from uncertain points. We make a distinction between outlier points and uncertain points in the sense that the uncertain points belong to the joint distribution of top-k classes but outside of each marginal distribution. Outlier points are outside of the joint distribution.

\subsubsection{Step-1: Detecting an Outlier point}
Let $N_{\pmb{kav}}(\pmb{\mu}_1, {\pmb{\sigma}_1}^2)$ and $N_{\pmb{kav}}(\pmb{\mu}_2, {\pmb{\sigma}_2}^2)$ be the $\pmb{kav}$ distributions of the top-2 classes respectively. Since $N_{\pmb{kav}}(\pmb{\mu}_1, {\pmb{\sigma}_1}^2)$ and $N_{\pmb{kav}}(\pmb{\mu}_2, {\pmb{\sigma}_2}^2)$ are independent Gaussians, their joint distribution is also Gaussian with parameters $\pmb{\mu}_{joint} = \pmb{\mu}_1 + \pmb{\mu}_2$ and $\pmb{\sigma}_{joint}^2 = {\pmb{\sigma}_1}^2 + {\pmb{\sigma}_2}^2$. Let us call this distribution $N_{\pmb{kav}-joint}(\pmb{\mu}_{joint}, {\pmb{\sigma}_{joint}}^2)$.

We define a test point as an outlier if its kernel activation vector is outside the $1^{st}-\textit{quantile}$ of joint squared mahalanobis distribution, $N_{MH-joint}(df, 2df)$ of top-2 classes. To achieve this, we first compute the Mahalanobis distance $d_{i,joint}$ from the \textbf{joint distribution of top-2 classes}
\begin{equation}\label{eq:5}
    \begin{split}
     d_{i,joint} &= \sqrt{\sum_{k=1}^{N} \frac{((\pmb{kav}_i)_k - (\pmb{\mu}_1 + \pmb{\mu}_2)_k)^2}{(\pmb{\sigma}_{1}^2 + \pmb{\sigma}_{2}^2)_k}}\\
    \end{split}
\end{equation}
and use the following $1^{st}-\textit{quantile}$ decision rule ($\leq \mu + \sigma$) to detect an outlier point:
\begin{equation}\label{eq:6}
    \begin{split}
     &\textit{if} \,\,\,\,  {d^2}_{i,joint} \leq df + \sqrt{2df} \longrightarrow \, \, outlier\\
     &\textit{else} \,\, \longrightarrow \textit{certain or uncertain}\\
    \end{split}
\end{equation}

\subsubsection{Step-2: Detecting Uncertain points}
We make a distinction between uncertain points and outlier points in the sense that the \textit{belongingness} of an uncertain point is within the joint distribution, $N_{\pmb{kav}-joint}(\cdot)$ of $\pmb{kav}$ but outside of each marginal distribution, $N_{\pmb{kav}-i}$ of the top-k prediction classes. An outlier point (for example random noise) is outside of such a joint distribution. To get a belongingness of a point (be it outlier or uncertain) to a multi-variate Gaussian distribution, we note that the squared Mahalanobis distances from such a distribution follow a $\chi^2$ distribution

For a test point, $\pmb{x}_i$, let $d_{i1}$ and $d_{i2}$ be mahalanobis distances respectively from the top-2 classes. If $d_{i1}$ and $d_{i2}$ are within $k\%$ of each other ($k$ learned empirically), then it is an uncertain point. If the point is not detected as outlier or uncertain, we tag that point as a certain point wrt the top-2 predictions. 

In order to increase the confidence score and make the certain and uncertain points more separable, we add a controlled noise along the direction of the gradient for every test point as follows, similar to previous methods \cite{lee2018simple, liang2017enhancing}:
\begin{equation}\label{eq:7}
    \begin{split}
     \pmb{x}_{updated} &= \pmb{x} + \epsilon \cdot \textit{sign}\left(\nabla_{\pmb{x}} f_{M}(\pmb{x})\right)  \\
    \end{split}
\end{equation}
where $f_{M}(\cdot)$ is a modified neural network estimator whose Softmax predictions are replaced by the proposed uncertainty layer.

\section{Experiments}
In this section, we investigate the effectiveness of the proposed method in detecting $\textit{uncertain}$ points within the data distribution (section[3.2]), understand class distributions overlap in the decision manifold (section 3.3), discuss robustness against digits rotation and plot uncertainty confidence with increasing image occlusion. All tests are done on a single GPU machine and using the pretrained network wherever possible, otherwise we fine-tuned networks for accuracy (Resnet34 on CIFAR10 and SVHN).

\subsection{New Baseline}
We define a new baseline that distinguishes between $uncertain$ and $out-of-distribution$ ($outlier$) points without sacrificing prediction performance. For demonstration purposes and to assess the proposed framework, we first establish that our method does not incur loss of prediction performance compared to recent methods such as \cite{lee2018simple, liang2017enhancing}. For this we compare out-of-distribution detection on TinyImageNet testset while using CIFAR10 testset as in-distribution data. We note that our method outperforms Softmax, Eucliden and gives the same performance as \cite{lee2018simple}. In addition, the prediction accuracy, using the decision thresholds in the proposed framework is at par with the previous works. Although fitting the proposed method on top of \cite{lee2018simple} would be an interesting test, we have left as a future investigation.

\begin{figure}[!htb]
\begin{center}
    \minipage{0.25\textwidth}
        \centering
        \includegraphics[width=\linewidth]{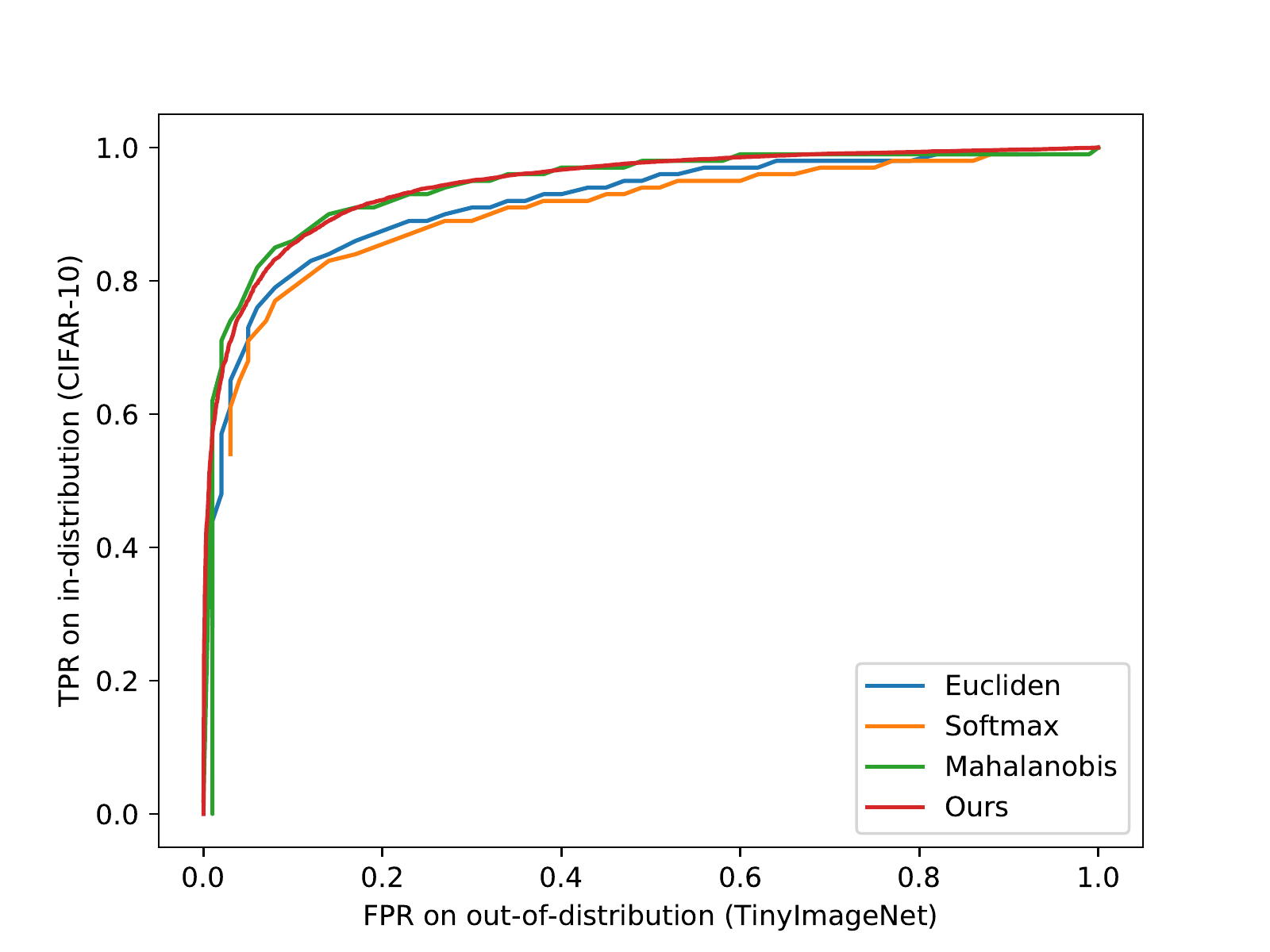}
        \captionof{subfigure}{CIFAR10}
    \endminipage
    \minipage{0.25\textwidth}
        \centering
        \includegraphics[width=\linewidth]{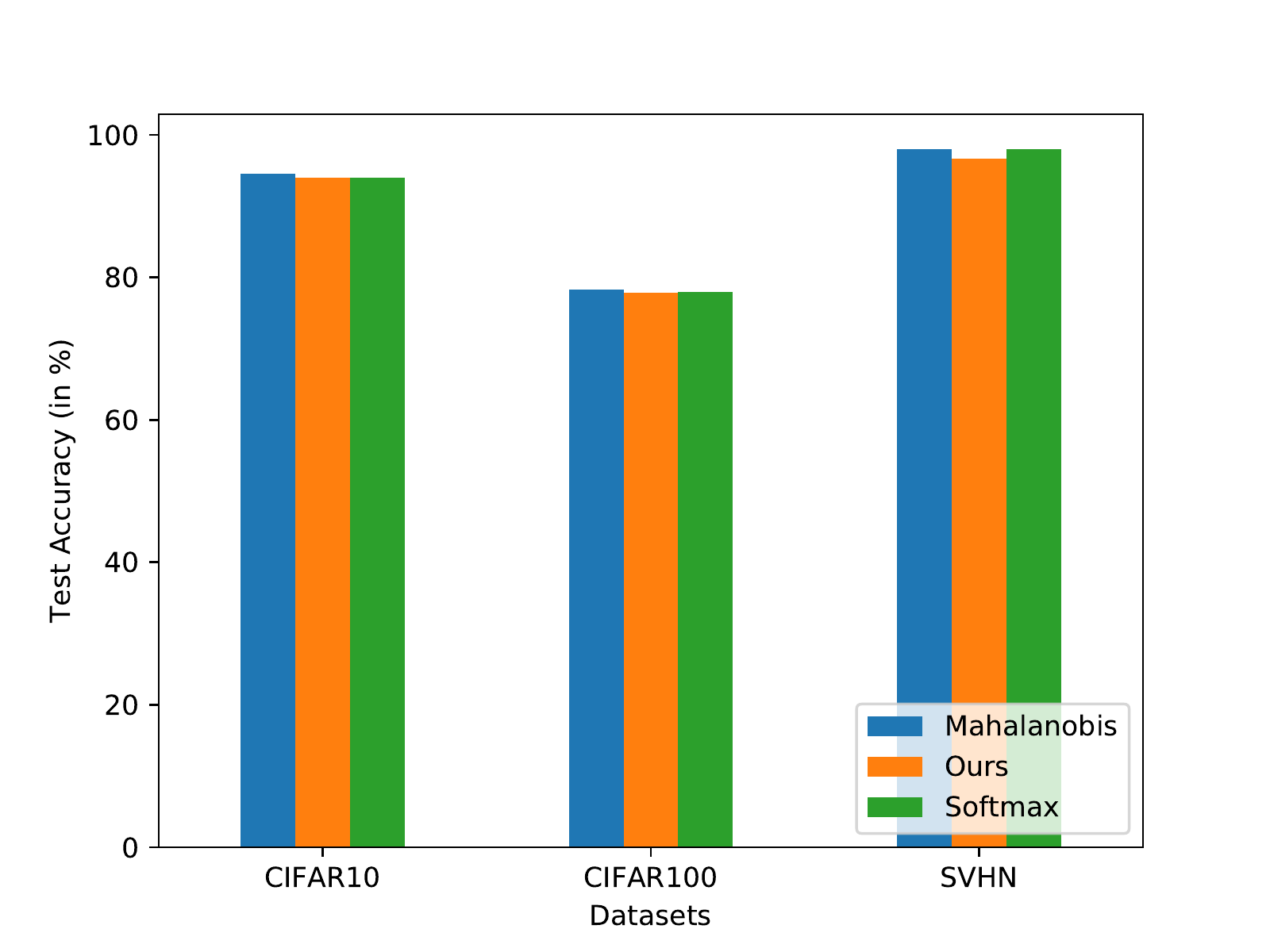}
        \captionof{subfigure}{CIFAR10}
    \endminipage
\caption{\textbf{(a)}. AUCROC curve of in-distribution CIFAR10 (True Positive Rate) versus the out-of-distribution TinyImageNet dataset (False Positive Rate) using ResNet34 finetuned for CIFAR10, \textbf{(b)}. Classification accuracy of CIFAR10 and SVHN on Softmax, \cite{lee2018simple} and the proposed method}.
\label{fig:fig2}
\end{center}
\end{figure}

\subsection{Detecting Uncertain points within data distribution}
With Fig.[\ref{fig:fig2}] as a benchmark, we now provide an assessment of the proposed method in detecting uncertain points that still remain in the data distribution. To generate uncertain points in the data distribution, we take a certain point and adversarially \cite{chakraborty2018adversarial} modify the image according to \cite{} until it just misses the true top-prediction. For example, an image of a plane under adversary could be classified as a bird.

\begin{figure}[!htb]
\begin{center}
    \minipage{0.25\textwidth}
        \centering
        \includegraphics[width=\linewidth]{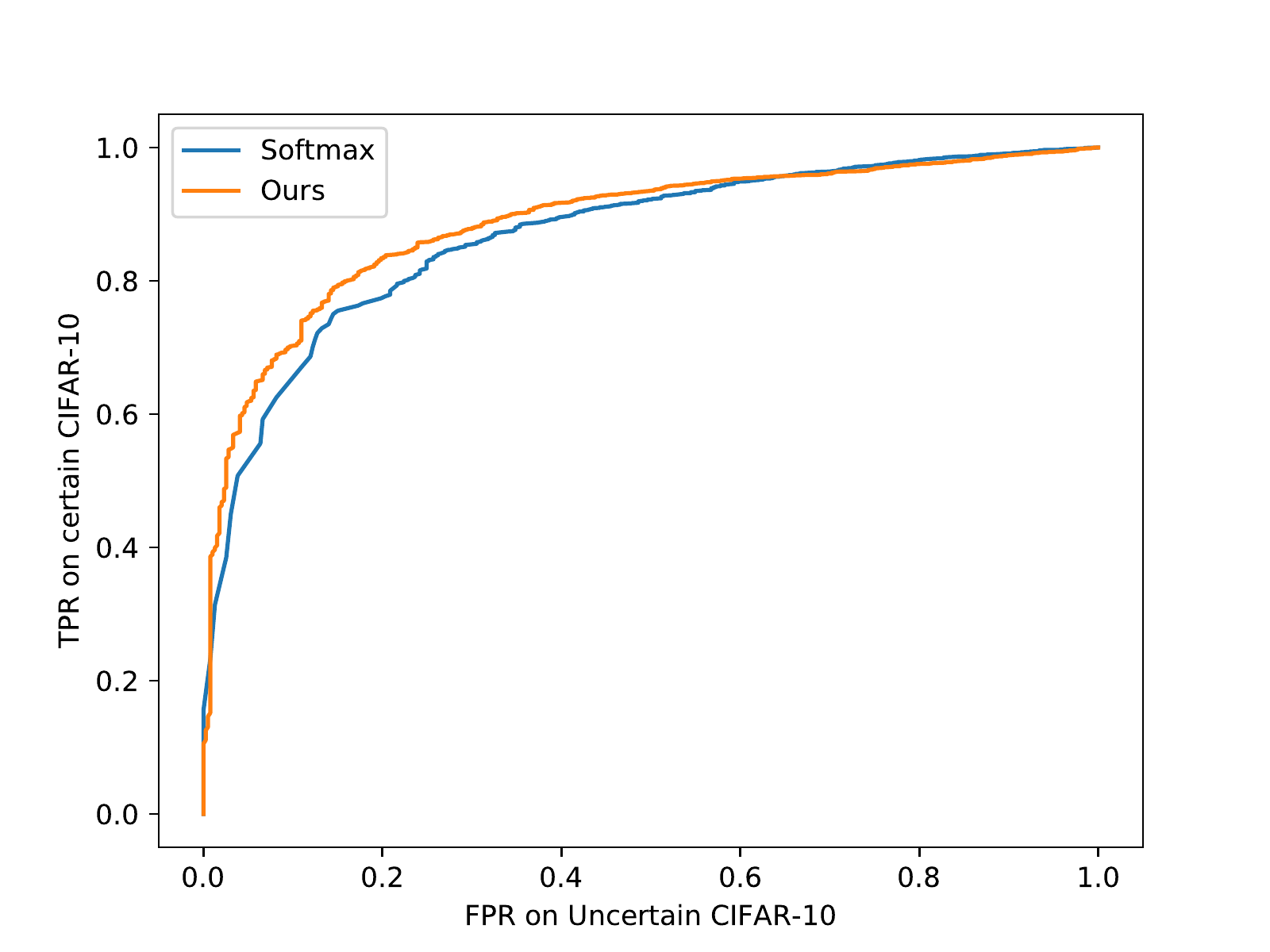}
        \captionof{subfigure}{ImageNet}
    \endminipage
    \minipage{0.25\textwidth}
        \centering
        \includegraphics[width=\linewidth]{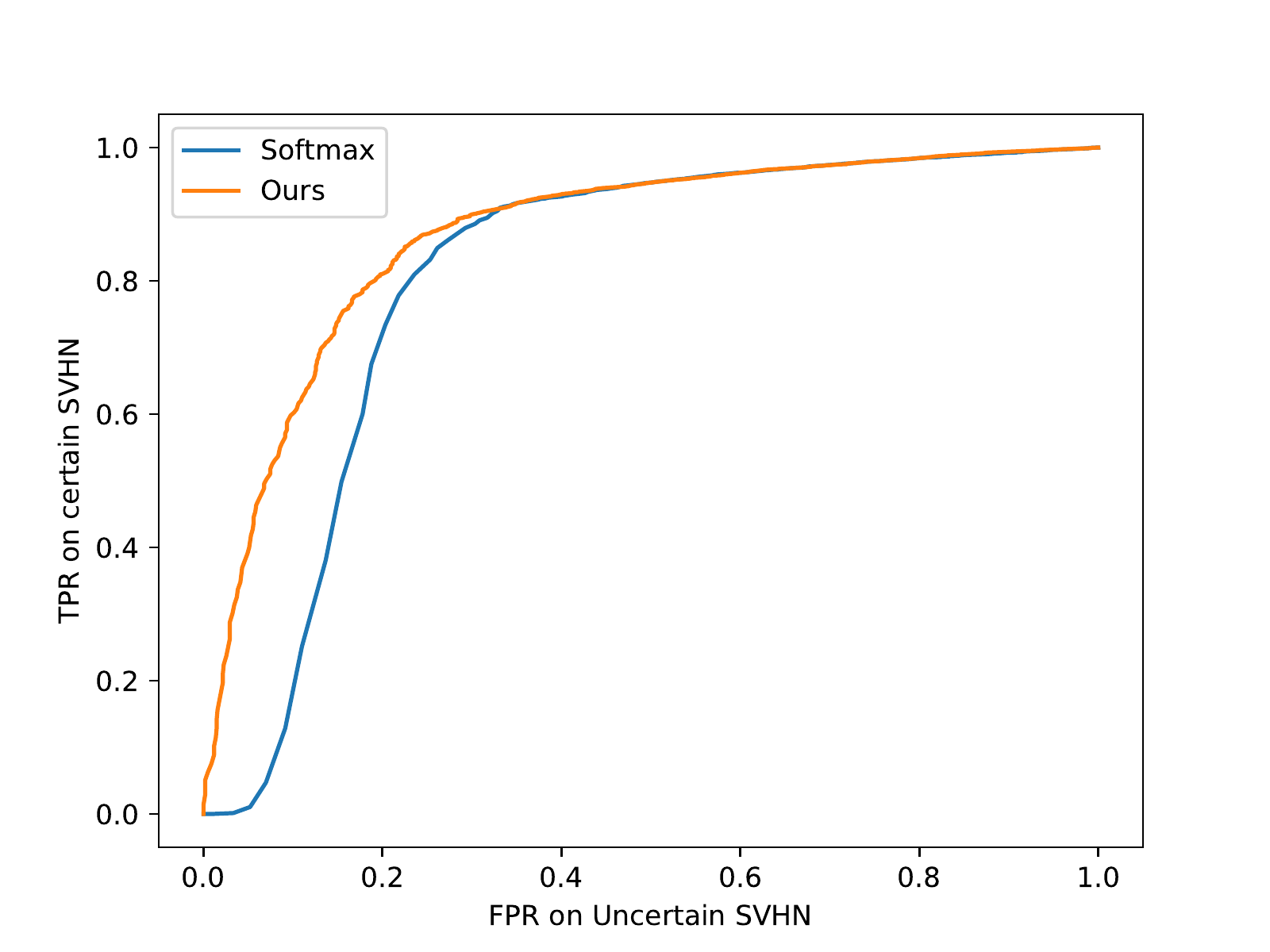}
        \captionof{subfigure}{CIFAR10}
    \endminipage
\caption{Baseline for uncertain image detection. Plot of TPR of in-distribution dataset against the FPR of (its adversarially generated) $\textit{uncertain}$ dataset.}
\label{fig:fig3}
\end{center}
\end{figure}

For assessment, we now have an in-distribution dataset containing images for which model is intended to provide correct predictions and an uncertain dataset for which model has confused the top-prediction. As before, we plot the AUCROC curve of TPR (in-distribution) versus FPR (uncertain) using the baseline proposed above and also compare the same with Softmax scores. As highlighted in Fig.[\ref{fig:fig3}], we see a significant performance increment compared to Softmax.

Next, to demonstrate that our finer grained uncertainty metric is able to make a distinction between uncertain and noise points, we take $10000$ random noise points and define this collection as an out of distribution dataset. The proposed method detects $\pmb{86\%}$ of the images as outlier - a new baseline for future works.

\subsection{Class Distribution Overlap}
\begin{figure*}[!htb]
\begin{center}
    \minipage{0.25\textwidth}
        \centering
        \includegraphics[width=\linewidth]{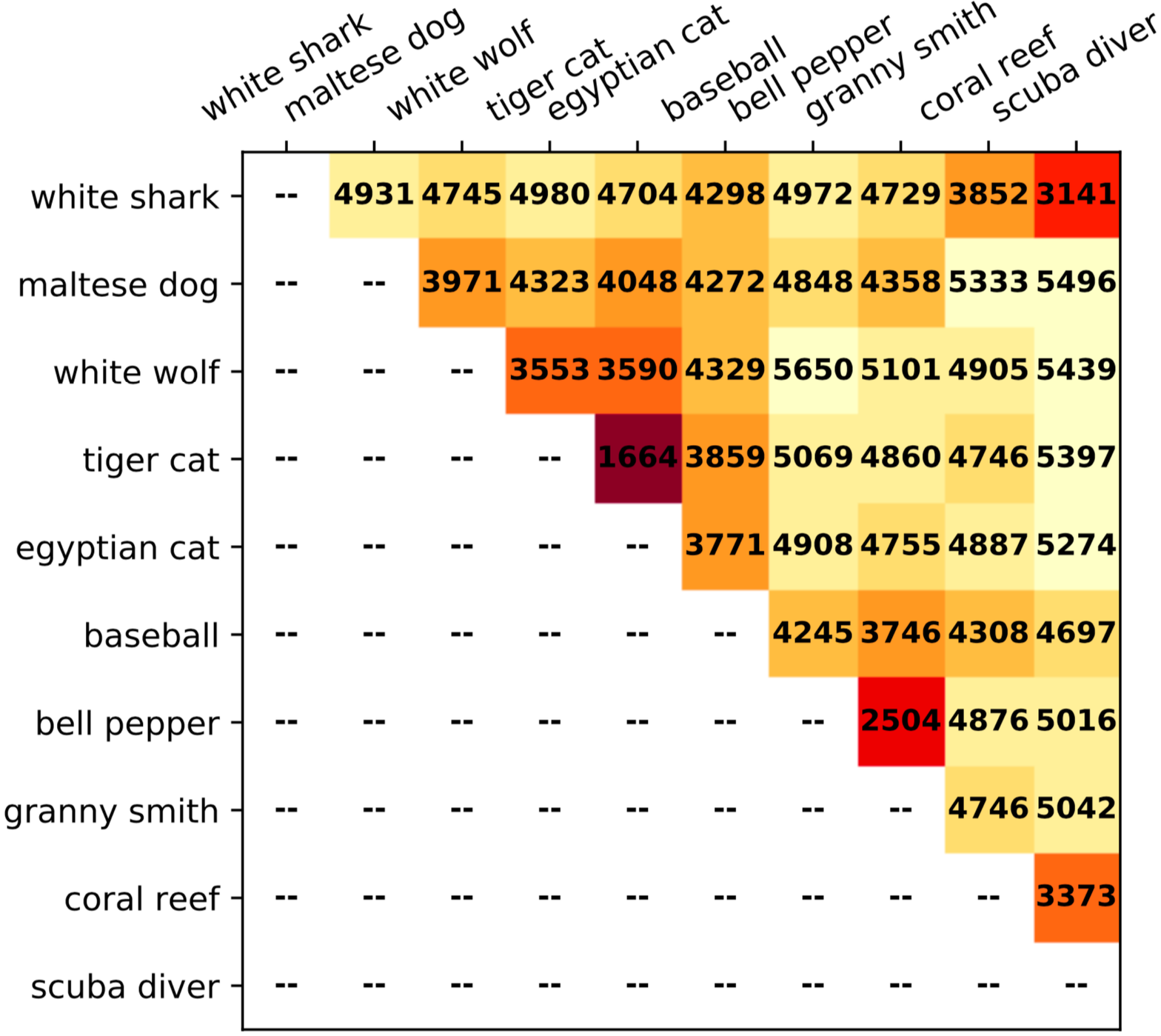}
        \captionof{subfigure}{ImageNet}
    \endminipage
    \minipage{0.25\textwidth}
        \centering
        \includegraphics[width=\linewidth]{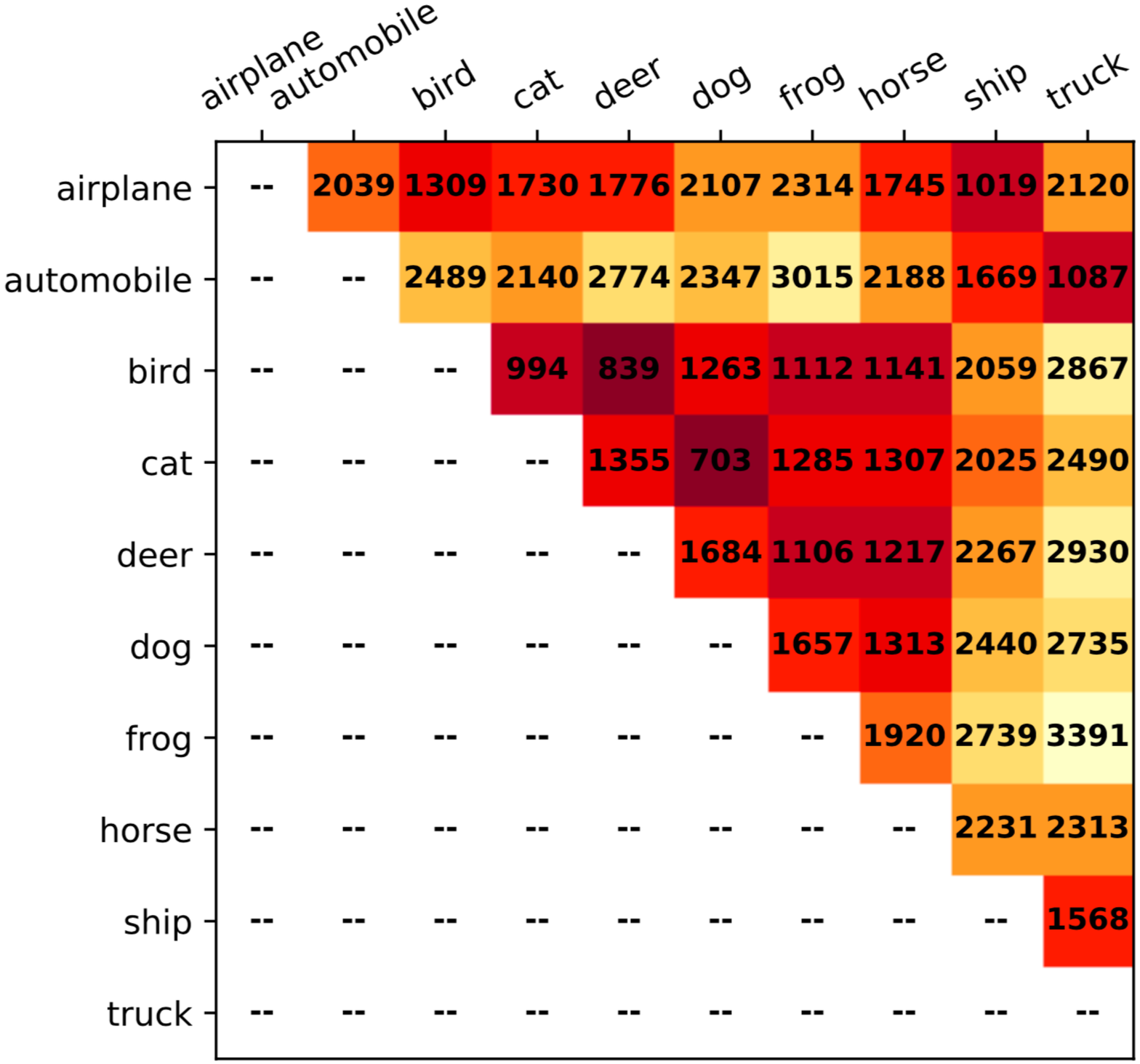}
        \captionof{subfigure}{CIFAR10}
    \endminipage
    \minipage{0.25\textwidth}
        \centering
        \includegraphics[width=\linewidth]{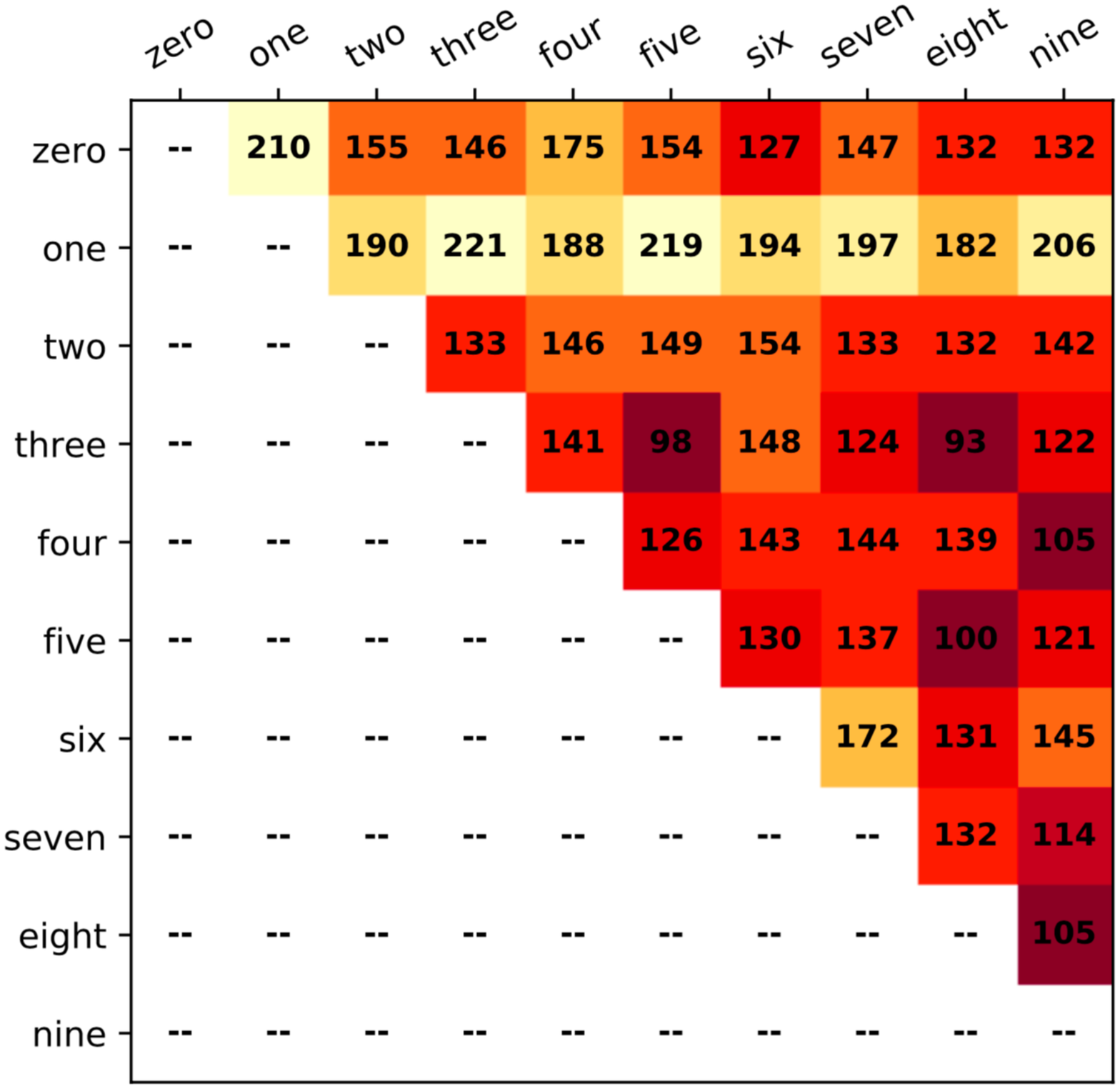}
        \captionof{subfigure}{MNIST}
    \endminipage
    \minipage{0.25\textwidth}
        \centering
        \includegraphics[width=\linewidth]{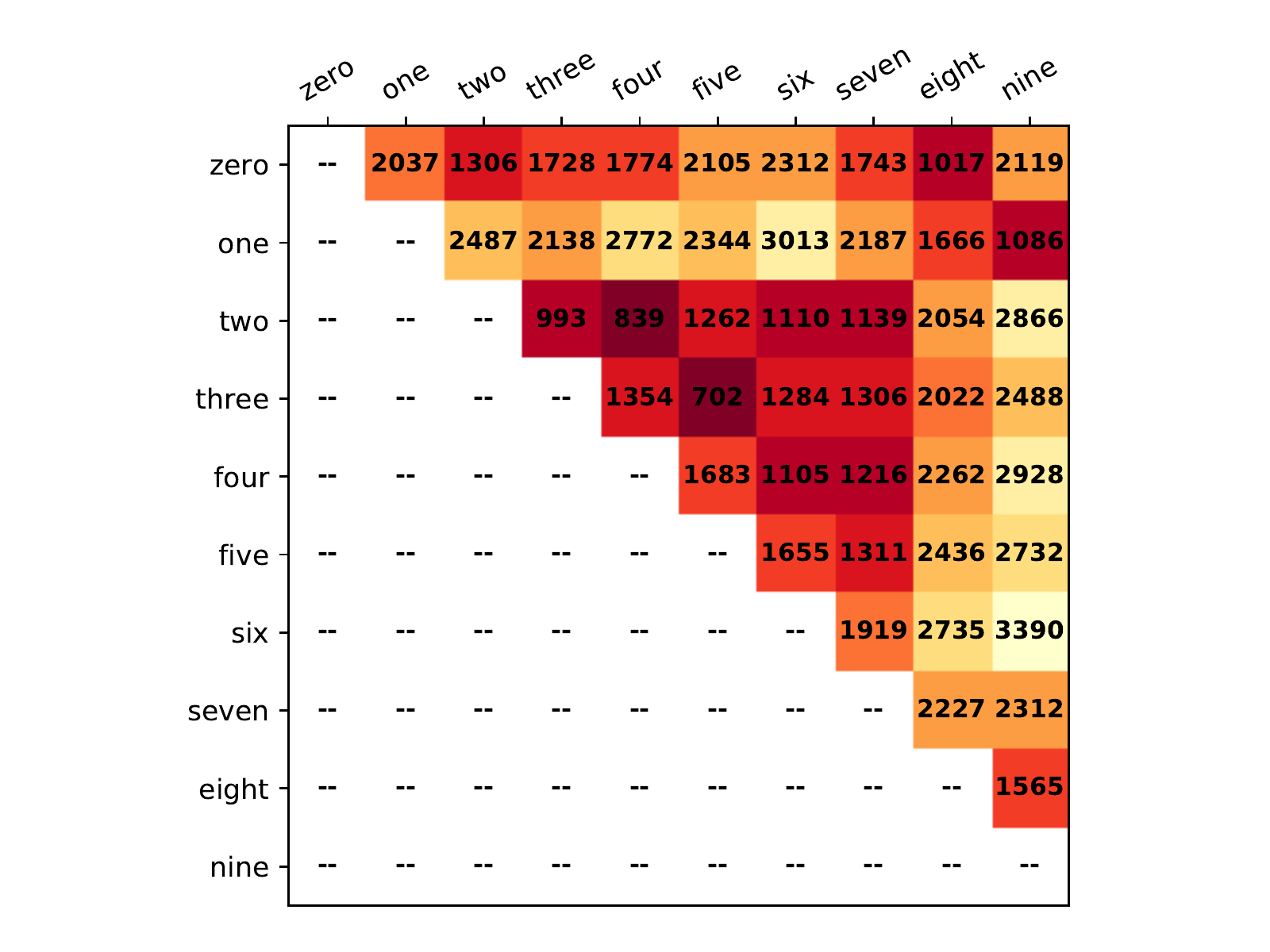}
        \captionof{subfigure}{SVHN}
    \endminipage    
\caption{Class distribution overlap for ImageNet (pretrained VGG16 \cite{simonyan2014very} and $10$ sampled classes), CIFAR10 (pretrained VGG16 with top-layer modified to $10$-classes and finetuned), MNIST (trained on LeNet \cite{lecun2015lenet}) and SVHN (pretrained VGG16 with top-layer modified to $10$-classes and finetuned). Numbers in each cell represent Bhattacharya distance where smaller number indicates higher overlap and vice versa.}
\label{fig:fig4}.
\end{center}
\end{figure*}

We quantify and investigate class distribution overlaps using the Bhattacharya distance as outlined in Section 2.1. One of the implications of this investigation is to understand if a Neural Network has learned a concept or is simply memorizing the training data. To substantiate our claim, we reason on the following two cases based on Fig.[\ref{fig:fig5}-(a)]. \textit{Case1}: we observe that the $\{``Egyptian \, Cat'', ``Tiger \, Cat''\}$ distribution pairs have a significantly high overlap (similarity) compared to any other pair. \textit{Case2}: we observe that the $\{``Scuba \, Diver'', ``Great \, White \, Shark''\}$ distribution pairs, although having high similarity compared to other pairs, are still significantly separated compared to \textit{Case1}. Next, in the ImageNet training data, many instances of $``Scuba \, Diver''$ class contain $``Great \, White \, Shark''$ in the image or one of its variant categories, (for example $``Hammerhead''$). This is not the case with $\{``Egyptian \, Cat'', ``Tiger \, Cat''\}$ pairs as their instances are fairly separated. This remark brings an obvious conclusion that the \textbf{model is learning concepts} (which is why $\{``Egyptian \, Cat'', ``Tiger \, Cat''\}$ pairs are close) \textbf{and not memorizing the training data} (otherwise $\{``Scuba \, Diver'', ``White \, Shark''\}$ overlap would be higher). Similar trends are observed in other distribution pairs in the 
ImageNet as well as other datasets (CIFAR10, MNIST, SVHN) as outlined in Fig.[\ref{fig:fig4}].

To the best of our knowledge, we believe that this finding is a unique contribution and has a profound implication in understanding the decision manifold and providing model interpretability.

\subsection{Image Rotation Test}
\begin{figure*}[!htb]
\begin{center}
    \minipage{0.25\textwidth}
        \centering
        \includegraphics[width=\linewidth]{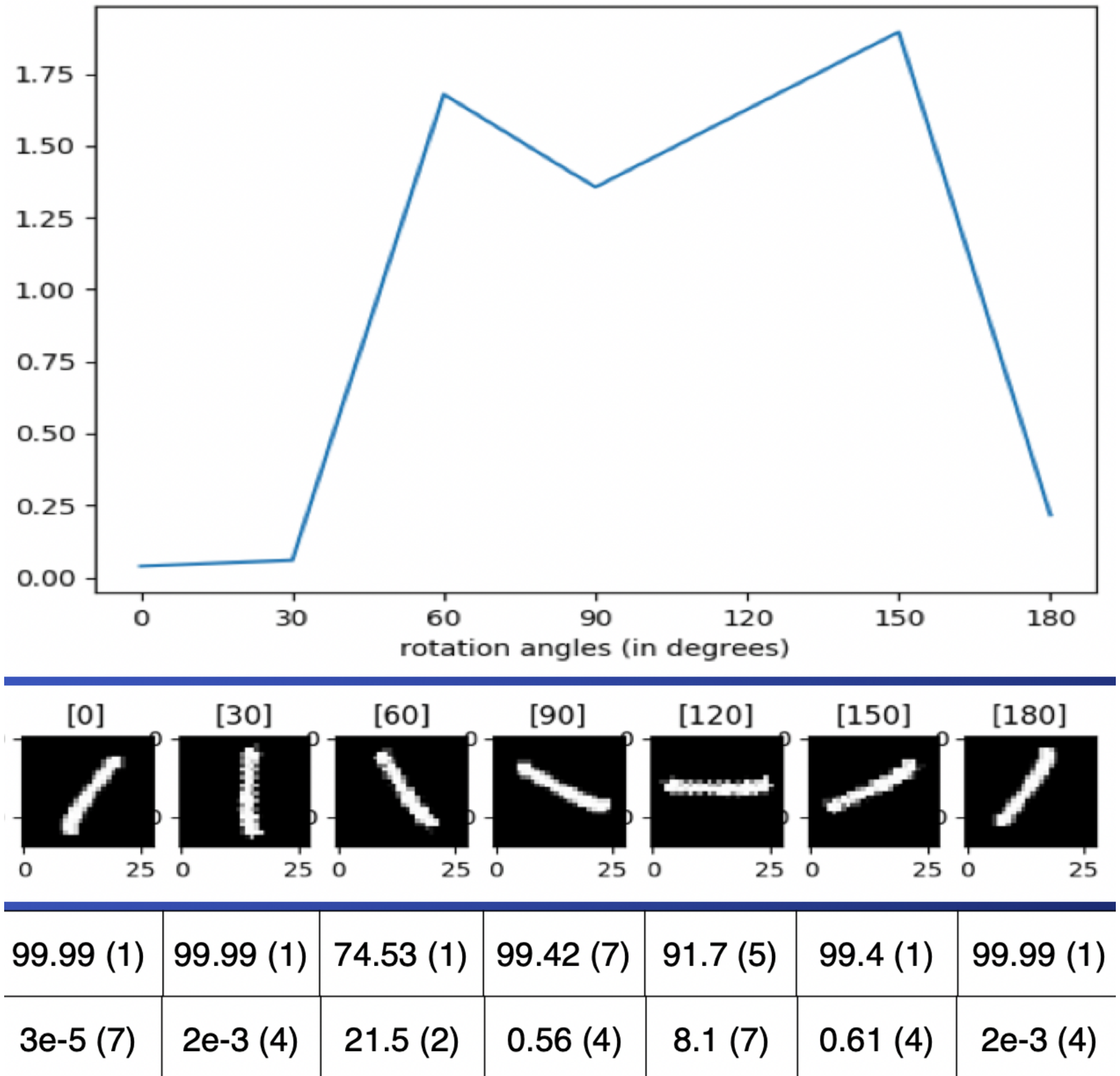}
        \captionof{subfigure}{Digit-1}
    \endminipage
    \minipage{0.25\textwidth}
        \centering
        \includegraphics[width=\linewidth]{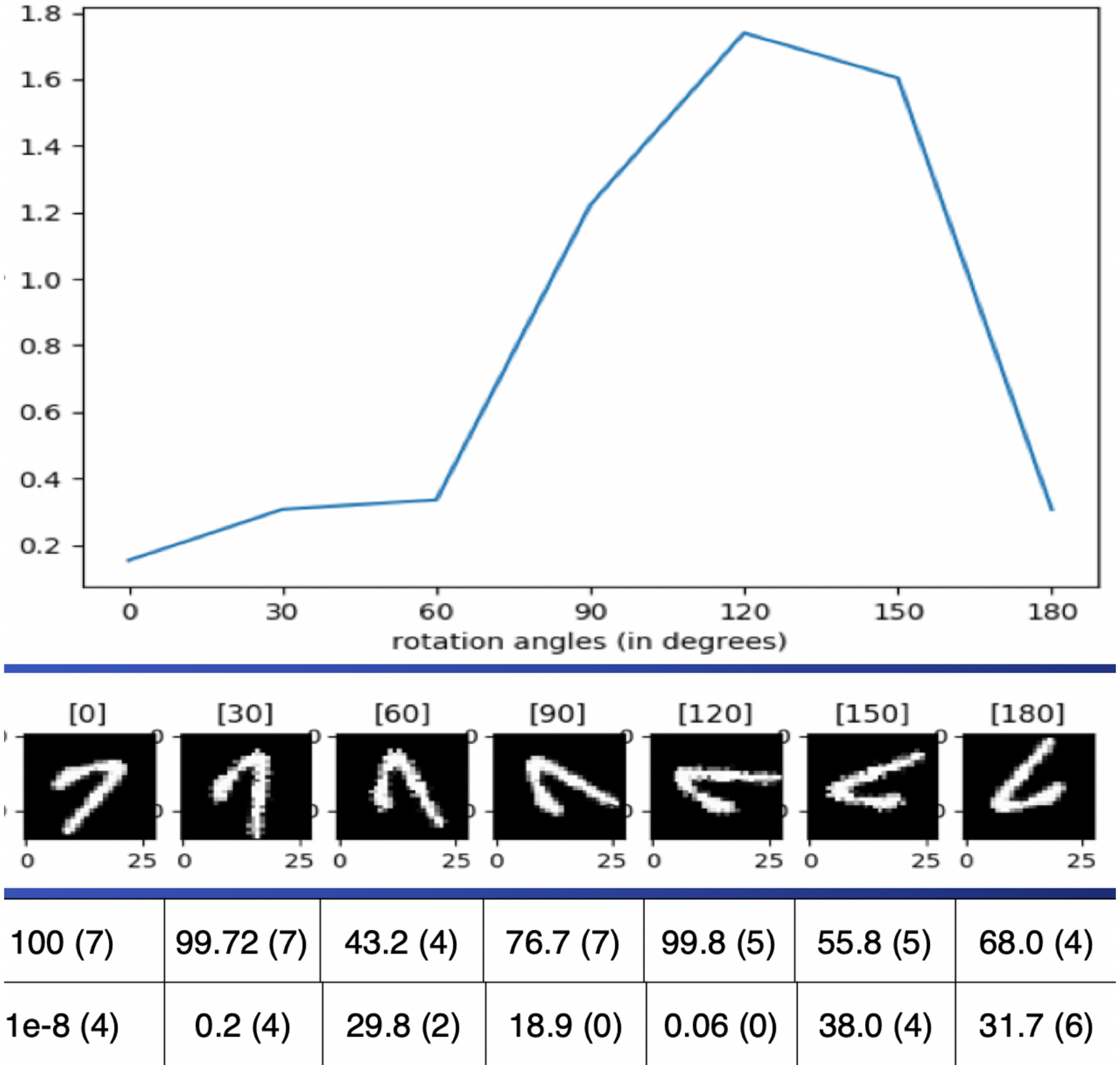}
        \captionof{subfigure}{Digit-7}
    \endminipage
    \minipage{0.25\textwidth}
        \centering
        \includegraphics[width=\linewidth]{images/rotate_one_curve.pdf}
        \captionof{subfigure}{Digit-6}
    \endminipage
    \minipage{0.25\textwidth}
        \centering
        \includegraphics[width=\linewidth]{images/rotate_seven_curve.pdf}
        \captionof{subfigure}{Digit-9}
    \endminipage    
\caption{Uncertainty score of the proposed method on MNIST image rotation using the LeNet architecture. The columns in the bottom of the figure represent top-2 softmax score along with prediction class in the parenthesis}.
\label{fig:fig5}.
\end{center}
\end{figure*}

We use the proposed approach to investigate the robustness of Neural networks confidence against image rotation similar to \cite{sensoy2018evidential} and compare against Softmax confidence. We observe in Fig[\ref{fig:fig5}] that a slight rotation of digit-1 leads to a significant drop is Softmax decision, (from $99\%$ at column-2 to $74\%$ in column-3). In addition, Softmax produces an overly confident score for confused prediction, for example, in Fig[\ref{fig:fig5}], digit-1 at $90\%$ rotation is detected as digit-7 with overly confident score of $99\%$, whereas the proposed method detects it as an highly uncertain point between digit-7 and digit-4. We conclude that the proposed method robustly detects uncertainty as the degree of rotation increases.

\subsection{Occlusion Test}
\begin{figure*}[!htb]
\begin{center}
    \minipage{0.25\textwidth}
        \centering
        \includegraphics[width=\linewidth]{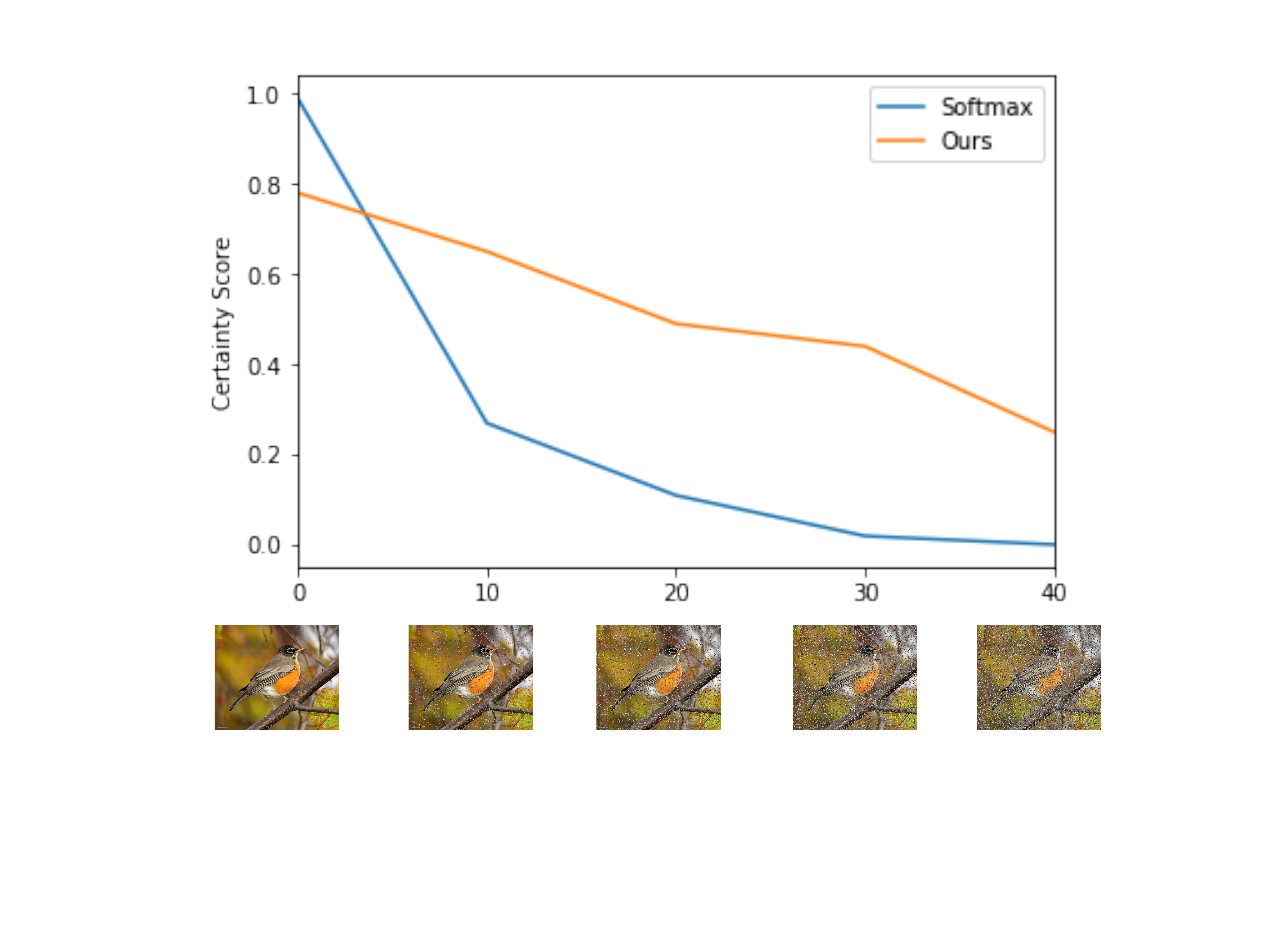}
        \captionof{subfigure}{Example-1}
    \endminipage
    \minipage{0.25\textwidth}
        \centering
        \includegraphics[width=\linewidth]{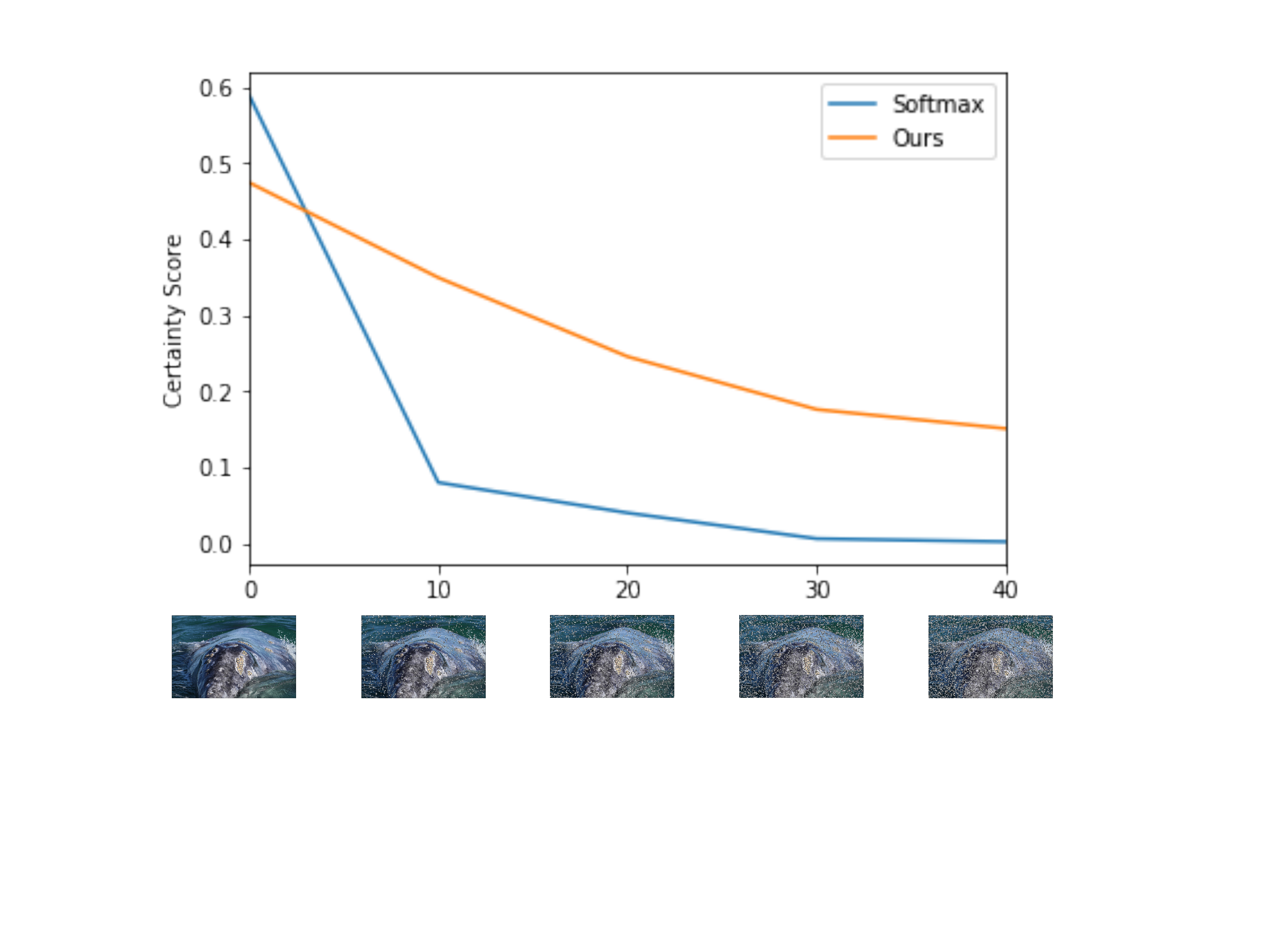}
        \captionof{subfigure}{Example-2}
    \endminipage
    \minipage{0.25\textwidth}
        \centering
        \includegraphics[width=\linewidth]{images/robin_uncertainty.pdf}
        \captionof{subfigure}{Example-3}
    \endminipage
    \minipage{0.25\textwidth}
        \centering
        \includegraphics[width=\linewidth]{images/hippopotamus_uncertainty.pdf}
        \captionof{subfigure}{Example-4}
    \endminipage    
\caption{Plot of uncertainty score with increasing salt and pepper noise. Images are taken from testset data in Table.[1]. We note that the Softmax confidence decays faster than our method for the fixed target prediction which makes our method robust to noise.}
\label{fig:fig6}
\end{center}
\end{figure*}

To quantify the robustness in confidence of the model prediction, we assess empirically, how the confidence degrades with increasing noise in the input space. For a given test image, we iteratively generate its noisy versions using the salt and pepper noise \cite{bovik2010handbook} with increasing degree of noise. In Fig.[\ref{fig:fig6}], we note that the proposed method is more confident about the target prediction compared to Softmax.

To perform a reliable experiment that covers the entire dataset, we compute occluded images (with increasing degree of noise) for every image in the test dataset and determine the average area under curve for occlusion. We compare the AUC of our method against Softmax, and as shown in Table.[\ref{tab:tab1}], our method significantly outperforms Softmax.

\begin{table*}[!htb]
\caption{Average AUC for the \textit{occluded} test-set constructed using the Salt and Pepper noise. Higher number is better.}
\label{tab:tab1}
\vskip 0.15in
\begin{center}
\begin{small}
\begin{sc}
\begin{tabular}{ccccccc}
\hline
 & \multicolumn{2}{c}{CIFAR10 (\%)} & \multicolumn{2}{c}{SVHN (\%)} & \multicolumn{2}{c}{Imagenet20 (\%)} \\ \hline
Noise (\%) & Softmax & Ours & Softmax & Ours & Softmax & Ours \\ \hline
0 & 97.03 & \multicolumn{1}{c|}{\textbf{97.52}} & 98.14 & \multicolumn{1}{c|}{\textbf{98.99}} & 89.63 & \textbf{91.57} \\
10 & 87.77 & \multicolumn{1}{c|}{\textbf{92.13}} & 89.13 & \multicolumn{1}{c|}{\textbf{94.56}} & 81.74 & \textbf{88.39} \\
20 & 71.02 & \multicolumn{1}{c|}{\textbf{80.94}} & 75.82 & \multicolumn{1}{c|}{\textbf{85.07}} & 69.58 & \textbf{74.92} \\
30 & 60.37 & \multicolumn{1}{c|}{\textbf{70.77}} & 63.71 & \multicolumn{1}{c|}{\textbf{73.82}} & 62.81 & \textbf{70.14} \\
40 & 57.81 & \multicolumn{1}{c|}{\textbf{65.89}} & 60.66 & \multicolumn{1}{c|}{\textbf{68.38}} & 55.24 & \textbf{61.33} \\
50 & 60.66 & \multicolumn{1}{c|}{\textbf{65.90}} & 59.83 & \multicolumn{1}{c|}{\textbf{67.81}} & 54.13 & \textbf{60.78}\\
\bottomrule
\end{tabular}
\end{sc}
\end{small}
\end{center}
\vskip -0.1in
\end{table*}

\section{Conclusion}
Proposed method presents a finer grained uncertainty measure that distinguishes outlier points from uncertain points. Our approach is based off of simple framework of kernal activations, is scalable to deep networks, is scalable to wider Softmax layer, sits on top of all neural networks and requires minimal disk storage. It is imperative to have confident model predictions in order to be able to use in critical real-world applications. We believe that our work is a meaningful contribution to achieve such confidence and prevent adversarial attacks robustly. In the future, we would like to investigate the effects of proposed method in multi-label classification tasks and in language modeling as well as kernel Mahalanobis distances \cite{haasdonk2009classification} to achieve prior separation is higher dimension manifold.

% \clearpage
% \bibliographystyle{plain}
% \bibliography{output.bbl}
\end{document}